\documentclass[runningheads]{llncs}
\usepackage{graphicx}
\usepackage{authblk}
\usepackage{graphicx}
\usepackage{subfigure}
\usepackage{array}
\usepackage{float}
\usepackage{booktabs}
\usepackage{bbding}
\usepackage{etoolbox}
\usepackage{color}
\usepackage{soul}
\usepackage{algorithm}
\usepackage{algorithmic}
\usepackage{cite}
\usepackage{balance}
\usepackage{amsmath, amssymb}
\usepackage{makecell}
\usepackage{multirow}
\usepackage{hyperref}
\usepackage[T1]{fontenc}
\usepackage{orcidlink}

\usepackage{graphicx,verbatim}

\begin{document}
\title{SurgSora: Object-Aware Diffusion Model for Controllable Surgical Video Generation}
\titlerunning{SurgSora}
%
\begin{comment}  %% Removed for anonymized MICCAI 2025 submission

\end{comment}

\author{Tong Chen\inst{1*}\orcidlink{ http://orcid.org/0000-0003-4312-7151} \and
Shuya Yang\inst{2*} \and
Junyi Wang\inst{3*}\and
Long Bai\inst{3\dagger}\orcidlink{https://orcid.org/0000-0002-9762-6821}\and
Hongliang Ren\inst{3}\orcidlink{https://orcid.org/0000-0002-6488-1551} \and
Luping Zhou\inst{1\dagger}\orcidlink{ http://orcid.org/0000-0002-6406-2505}
}
\authorrunning{T. Chen et al.}

\institute{The University of Sydney, Sydney, Australia
\and The University of Hong Kong, Hong Kong SAR, China
\and The Chinese University of Hong Kong, Hong Kong SAR, China\\
\email{tche2095@uni.sydney.edu.au, b.long@ieee.org, luping.zhou@sydney.edu.au}}
\maketitle              % typeset the header of the contribution

\begin{abstract}
Surgical video generation can enhance medical education and research, but existing methods lack fine-grained motion control and realism. We introduce SurgSora, a framework that generates high-fidelity, motion-controllable surgical videos from a single input frame and user-specified motion cues. Unlike prior approaches that treat objects indiscriminately or rely on ground-truth segmentation masks, SurgSora leverages self-predicted object features and depth information to refine RGB appearance and optical flow for precise video synthesis. It consists of three key modules: (1) the Dual Semantic Injector, which extracts object-specific RGB-D features and segmentation cues to enhance spatial representations; (2) the Decoupled Flow Mapper, which fuses multi-scale optical flow with semantic features for realistic motion dynamics; and (3) the Trajectory Controller, which estimates sparse optical flow and enables user-guided object movement. By conditioning these enriched features within the Stable Video Diffusion, SurgSora achieves state-of-the-art visual authenticity and controllability in advancing surgical video synthesis, as demonstrated by extensive quantitative and qualitative comparisons. Our human evaluation in collaboration with expert surgeons further demonstrates the high realism of SurgSora-generated videos, highlighting the potential of our method for surgical training and education. Our project is available at \href{https://surgsora.github.io/}{\textcolor{magenta}{surgsora.github.io}}.

\end{abstract}
\renewcommand{\thefootnote}{}
\footnotetext{* Equal Contribution; \ \inst{\dagger} Corresponding Author.}
\section{Introduction}
Surgical video generation has the potential to enhance medical education, clinician training, and AI-driven surgical analysis by providing realistic and controllable visual representations of complex procedures~\cite{cho2024surgen,sun2024bora,ozawa2021synthetic}. However, existing methods face two key challenges: visual authenticity and generation controllability. Different strategies have been proposed to generate high-quality endoscopic videos. 
Endora~\cite{li2024endora} is an endoscopy simulator capable of replicating diverse endoscopic scenarios while lacking precise controlling abilities. MedSora~\cite{wang2024optical} focuses on past temporal coherence to achieve high-fidelity forward video generation, while predicting based on the optical flow of past frames is challenging due to the complexity of surgical scenaris.
Moreover, most current models indiscriminately process entire scenes, failing to differentiate individual objects, leading to blurred boundaries and unrealistic motion dynamics. Iliash et al.~\cite{iliash2024interactive} attempt to improve object awareness using segmentation masks, while this approach requires additional object-level annotations and hard object boundaries, limiting the ability to model smooth transitions and fine anatomical details. 

Beyond realism, precise motion control is critical for replicating surgical actions such as tool manipulation and tissue interaction~\cite{wang2024controllable}. However, most existing methods lack this capability, making it difficult to generate videos that follow user-specified actions. In the context of general video generation, text prompts are a commonly used control method~\cite{li2018video, wang2024recipe,cho2024surgen,sun2024bora,iliash2024interactive}. However, surgical scenarios often involve precise and detailed actions, which are difficult to fully capture through textual descriptions. This leads to a domain gap between language and motion, making it challenging for text-based guidance to express fine-grained surgical actions. In addition, surgical triplets (tissue-instrument-action) have also been used for video generation tasks~\cite{nwoye2025surgical,nwoye2022rendezvous,yeganeh2024visage}, while triplets do not include specific motion and directional information, making it difficult to comprehensively represent the rich information contained in surgical videos. 

To this end, we propose SurgSora, an object-aware trajectory-controlled RGB-Depth Flow Diffusion model that introduces self-predicted object-relevant RGB and depth features for precise object localization — eliminating the need for segmentation masks. Instead of relying on text-based control, SurgSora utilizes motion trajectories, offering fine-grained, user-directed control over object movements, enabling the generation of realistic and clinically relevant surgical videos. We can freely generate realistic videos by providing motion guidance using our SurgSora, offering abundant surgical training and education materials.

Our contribution can be summarized as: \underline{\textbf{(i)}} We present the first work on motion-controllable surgical video generation using a diffusion model, which allows fine-grained control (both direction and magnitude) over the motion of surgical instruments and tissues, guided by intuitive motion cues provided by simple clicks.  \underline{\textbf{(ii)}} We propose the Dual Semantic Injector (DSI), which integrates object-aware RGB-D semantic understanding. The DSI combines appearance (RGB) and depth information to better discriminate objects and capture complex anatomical structures, providing an accurate representation of the surgical scene.  \underline{\textbf{(iii)}} We introduce the Decoupled Flow Mapper (DFM), which effectively fuses optical flow with semantic-RGB-D features at multiple scales. This fusion serves as the guidance conditions for a frozen Stable Video Diffusion model to generate realistic surgical video sequences.  \underline{\textbf{(iv)}} Extensive experiments on the CoPESD dataset demonstrate the effectiveness of SurgSora in generating high-quality, motion-controllable surgical videos. We further involve surgeons verifying the authenticity of our videos to justify their usefulness in medical use. 

\section{Methodology}
\begin{figure}[htbp]
    \centering
    \includegraphics[width=0.95\linewidth, trim=0 15 0 30]{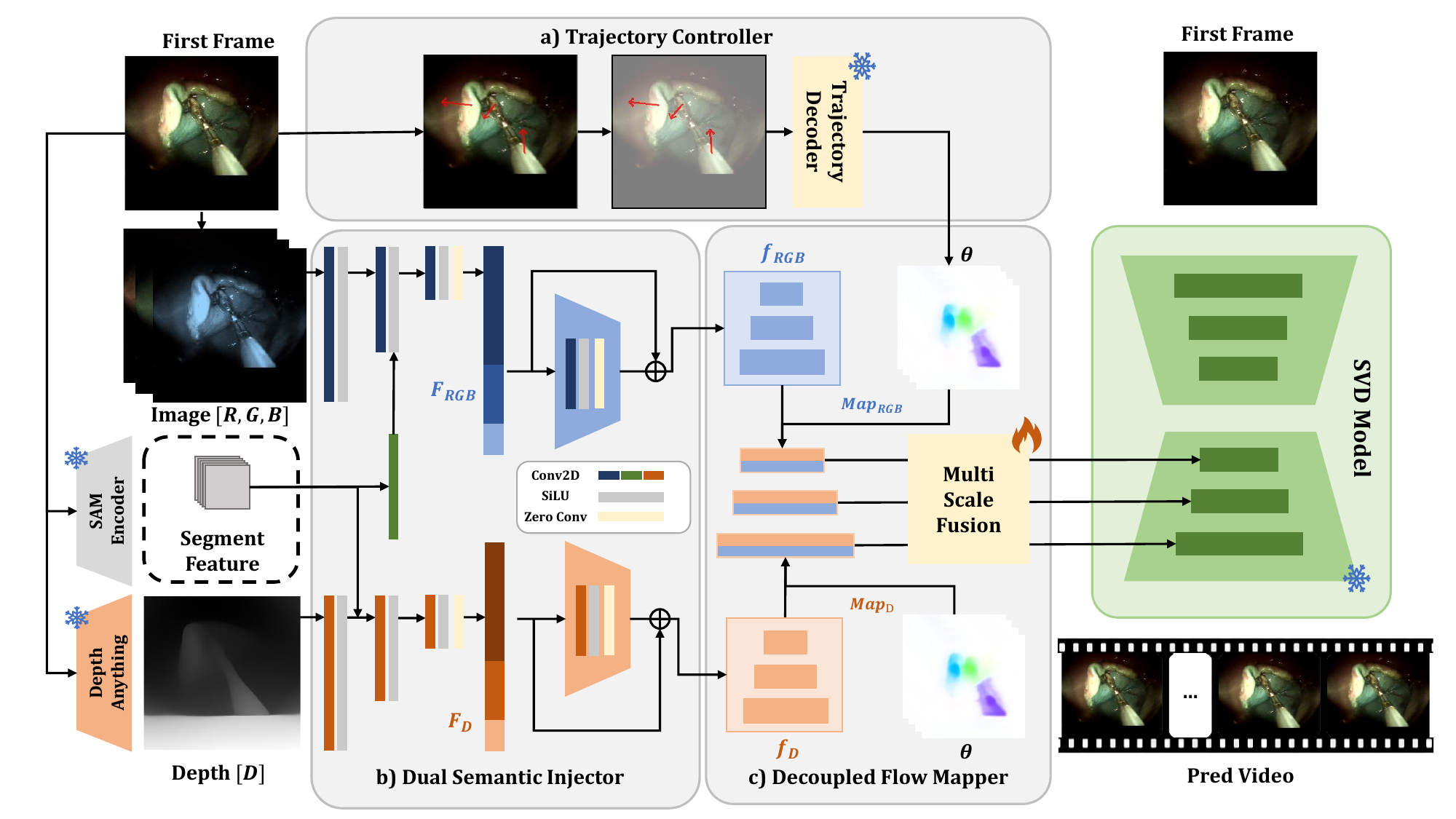}
    \caption{\textbf{SurgSora Pipeline:} \textbf{a)} Trajectory Controller module encodes trajectories into sparse optical flow. \textbf{b)} Dual Semantic Injector merges RGB and depth features with segment features separately. \textbf{c)} Decoupled Flow Mapper maps RGB and depth features into optical flow, then sends to Multi-Scale Fusion Block as condition.}
    \label{fig: Schematic}
\end{figure}
Our SurgSora framework, as illustrated in Fig.~\ref{fig: Schematic}, comprises three key modules: the Dual Semantic Injector (DSI) introduced in Sec.~\ref{sec: Segment Injector}, the Decoupled Flow Mapper (DFM) described in Sec.~\ref{sec: Decoupled Projector}, and the Trajectory Controller (TC) module detailed in Sec.~\ref{sec: Trajectory Controller}. Our model takes the first image frame $I_{RGB} \in \mathbb{R}^{3\times H\times W}$ as input. Based on $I_{RGB}$,  the corresponding segmentation features $f_{seg}$ and depth image $I_{D} \in \mathbb{D}^{1\times H\times W}$ are generated from the pretrained Segment Anything Model~\cite{kirillov2023segment} and the Depth Anything V2~\cite{yang2024depth}. The segment feature is injected into the RGB and Depth features in the DSI module to extract object-aware image features $f^r_{RGB}$ and depth features $f^r_{D}$ at multi-scales $r$.
These features are then processed in the DFM module, where the optical flow $\theta \in \mathbb{O}^{(T-1)\times2 \times H\times W}$ (with $T$ as the total number of frames of the generated video), is resized and used to transform $f_{RGB}^{r}$ and $f_{D}^{r}$ independently. The transformed features are fused using the Multi-Scale Fusion (MSF) Block at different scales. These multi-scale fused features are then used as conditions for a frozen Stable Video Diffusion (SVD) model to generate the video.

\subsection{Dual Semantic Injector}
\label{sec: Segment Injector}
Traditional methodologies primarily rely on RGB images as input to create dynamic visual content. 
\textcolor{black}{While effective in certain applications, this approach suffers from significant limitations in depth perception and scene understanding.}
Specifically, relying solely on RGB data complicates accurately capturing spatial relationships between objects, leading to deficiencies in visual coherence and object segmentation in generated videos. 
\textcolor{black}{To address these challenges, we introduce the Dual Semantic Injector (DSI) module, a dual-branch architecture that enhances object awareness by integrating segmentation features into both the RGB and depth feature branches. Unlike traditional methods that depend solely on RGB images, we estimate and incorporate a depth map to provide crucial geometric cues. These cues improve the understanding of spatial relationships between objects and overall scene structure, making it especially beneficial for complex tasks like surgical video synthesis. Furthermore, to better discriminate between objects, object segmentation is leveraged to refine both RGB and depth features.}
The segment features $f_{seg}$ are combined with RGB images $I_{RGB}$ and depth images $I_{D}$ by passing through two separate processors $\phi_{RGB}$ and $\phi_{D}$ for feature extraction and fusion, followed by two separate encoders for further encoding. The Dual Semantic Injector can be formulated as:
\begin{equation}
    f^r=\left\{
    \begin{aligned}
    &\boldsymbol{{Encoder}_{RGB}^r}(\boldsymbol{\phi_{RGB}}(I_{RGB},  f_{seg})), or      \\
    &\boldsymbol{{Encoder}_{D}^r}(\boldsymbol{\phi_{D}}(I_{D}, f_{seg}))  .
    \end{aligned}
    \right.
    \label{equ:encoder}
\end{equation} 
Recall that the superscript $r$ indicates different scales of feature maps extracted by the encoders. This dual encoding method synchronizes and harmonizes the enhanced features from RGB and depth channels to optimize the overall representation. \textcolor{black}{The segmentation features enhance the semantic understanding compared with using the original RGB and depth features, significantly improving the discrimination of foreground and background, enhancing depth estimation, and ultimately contributing to more realistic and referenceable video predictions.}

\subsection{RGB-Depth Frame Mapper} 
\label{sec: Decoupled Projector} 
Recent studies~\cite{zhang2023adding,niu2024mofa, rombach2022high} have highlighted the advantages of integrating additional latent space information into diffusion models to enhance their output quality. Building on this insight, our Decoupled Flow Mapper (DFM) module integrates spatial and temporal data from image and optical features to generate sequential videos effectively. The DFM uses object-aware RGB and depth features from the Dual Semantic Injector (DSI) module, transformed spatially by resized optical flow to produce enriched video sequences.

Specifically, feature maps $f^r \in \mathbb{R}^{C_r\times H_r\times W_r}$ from the DSI module are modified according to the resized optical flow $\theta^r \in \mathbb{O'}^{(T-1)\times2 \times H_r\times W_r}$, aligning them spatially for each frame. The features are shifted by $\mathrm{d}x$ and $\mathrm{d}y$, the displacements provided by $\theta^r_t$, with new positions computed as $x' = x + \mathrm{d}x$ and $y' = y + \mathrm{d}y$. Interpolation determines the new pixel values at these coordinates, effectively merging the depth information's structural insights with the RGB data's textural details. This decoupled approach ensures that each feature type is optimally utilized, enhancing the video generation process. The fused features from separate streams are then passed the \textit{Multi-Scale Fusion Block} (MSF) for final video output. The MSF block shall fuse the optical-flow-transformed RGB and depth features by concatenating them at different scales and then fusing them with 3D convolution blocks and an activation block, which can be formulated as: 
\begin{equation}
    \check{f}_{fuse}^r = \mbox{SiLU}(\mbox{Conv3d}(\mbox{Conv3d}(\mbox{ConCAT}(\hat{f}^r_{RGB}, \hat{f}^r_{D})))).
\end{equation}
The fused feature $\check{f}_{fuse}^r$ is then used to assist a pretrained Stable Video Diffusion (SVD) Model for conditional video generation.
The integration of optical flow information substantially enhances the temporal continuity and smoothness for generation, enabling accurate capture of scenes and complex objective motions. By utilizing fused data, the model gains a richer understanding of scene depth and structure, ensuring visual authenticity and adaptability to nuanced changes. This approach improves video quality through diversity-aware fusion, providing a dynamic and precise scene representation.

\subsection{Trajectory Controller}
\label{sec: Trajectory Controller}
Surgical videos demand higher precision compared to natural scene videos, particularly because generating them from a single image introduces significant challenges and reduces referential accuracy. Due to this we employ trajectory for precision control, which employs a pre-trained trajectory decoder from \cite{zhan2019self}. The surgeon inputs the first frame image and then clicks to set trajectories. The trajectories and image will be encoded separately, concatenated together, and then decoded by the trajectory decoder into optical flows as a condition to guide the following generation. By involving the TC module, generated videos will be more referencable and convenient for generating customized surgical videos.

\section{Experiment}

\textbf{Dataset and Implementation Details.}
We utilize the publicly available CoPESD dataset~\cite{wang2024copesd}, which was collected from 20 videos using both conventional endoscopic submucosal dissection (ESD) and the DREAMS system~\cite{gao2024transendoscopic}, performed on in-vivo porcine models. 
The videos were recorded at a frame rate of 30 Hz with an original resolution of 1920 $\times$ 1080, which was cropped to 1300 $\times$ 1024. After an expert surgeon provided temporal annotations of ESD activities, video segments corresponding to submucosal dissection were separately extracted. We extract 21-frame video clips from the datasets, then resize them into $256\times197$ resolution and pad them to $256\times256$ resolution as the training and testing set.
for training. For the training process, we use the AdamW~\cite{loshchilov2017decoupled} with a learning rate of $2\times10^{-5}$, a batch size of 4, and train on two A6000 GPUs.
\\
\\
\noindent \textbf{Comparison Methods and Evaluation Metrics.}
We assess the performance of our model against leading video generation models: the unconditional models Endora~\cite{li2024endora} and StyleGAN-V~\cite{skorokhodov2022stylegan}; and the conditional model MOFA-Video~\cite{niu2024mofa}.
For video quality evaluation, we employ metrics such as \textcolor{black}{Fréchet Video Distance (FVD)\cite{unterthiner2018towards} and Content-Debiased Fréchet Video Distance (CD-FVD)\cite{ge2024content} to assess temporal consistency and video realism. Frame realism and diversity are evaluated using Fréchet Inception Distance (FID)\cite{heusel2017gans} and Inception Score (IS)\cite{barratt2018note}.} Frame consistency is measured by CLIP cosine similarity~\cite{radford2021clip} between consecutive frames. For conditional generation, we employ PSNR~\cite{huynh2008scope} and SSIM~\cite{wang2004image} for content and structural accuracy, alongside optical flow metrics F1-epe and F1-all, to verify flow consistency with the input trajectories.

\subsection{Experimental Results}
\begin{table}[htbp]
	\caption{
 Quantitative Comparisons on CoPESD Dataset~\cite{wang2024copesd}.}
 	\centering
	\label{table:baseline}  
\resizebox{\textwidth}{!}{	
\begin{tabular}{c|c|c|cccccc|cc}
\hline
\multirow{2}{*}{Models} & \multirow{2}{*}{\makecell{Discriminate\\Avg Acc}}& \multirow{2}{*}{\makecell{Frame\\Consistency}$\uparrow$} &\multicolumn{6}{c|}{Video}  &\multicolumn{2}{c}{Optical-Flow}\\ \cline{4-11}
 & & &FVD$\downarrow$ &CD-FVD$\downarrow$ & FID$\downarrow$ & IS $\uparrow$  &PSNR $\uparrow$ &SSIM $\uparrow$ & F1-epe$\downarrow$ & F1-all$\downarrow$\\ \hline 
Endora~\cite{li2024endora}           &--     & 97.51\%  &1146.54 &1289.59 &205.93 & 2.239&--&-- &--&--\\
StyleGAN-V~\cite{digan}              &--  &98.02\% &857.16 &980.11 &166.03 &2.267&--&--&--&--\\
MOFA-Video~\cite{niu2024mofa}        &--       & 95.59\% &671.66 &692.44 & 96.31 & 2.685& 19.06 & 48.28\% &0.2620 & 265.54 \\
\textbf{SurgSora (Ours)} &53.33\%    & \textbf{98.70\%} & \textbf{395.65} & \textbf{535.95}  & \textbf{87.94} & \textbf{3.278} & \textbf{20.71}&\textbf{55.94\%} & \textbf{0.1477}  & \textbf{149.89}\\ \hline
\end{tabular}}
\end{table}

\begin{figure}[]
    \centering
    \includegraphics[width=1\linewidth, trim=-10 200 0 95]{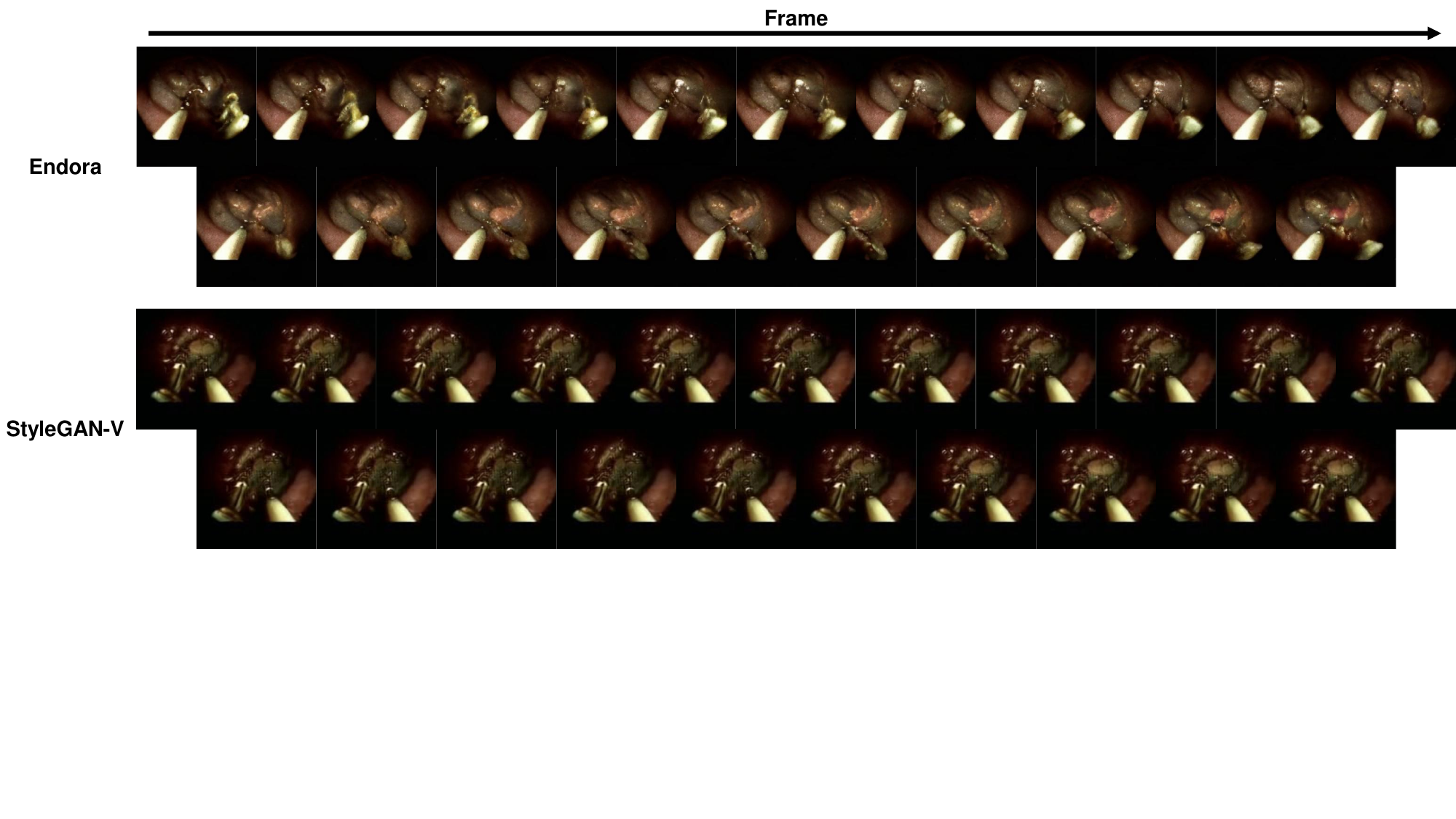}
    \caption{Comparison of unconditional results on CoPESD dataset.}
    \label{fig: uncon-baseline}
\end{figure}

\begin{figure}[t]
    \centering
    \includegraphics[width=1\linewidth, trim=-10 30 20 0]{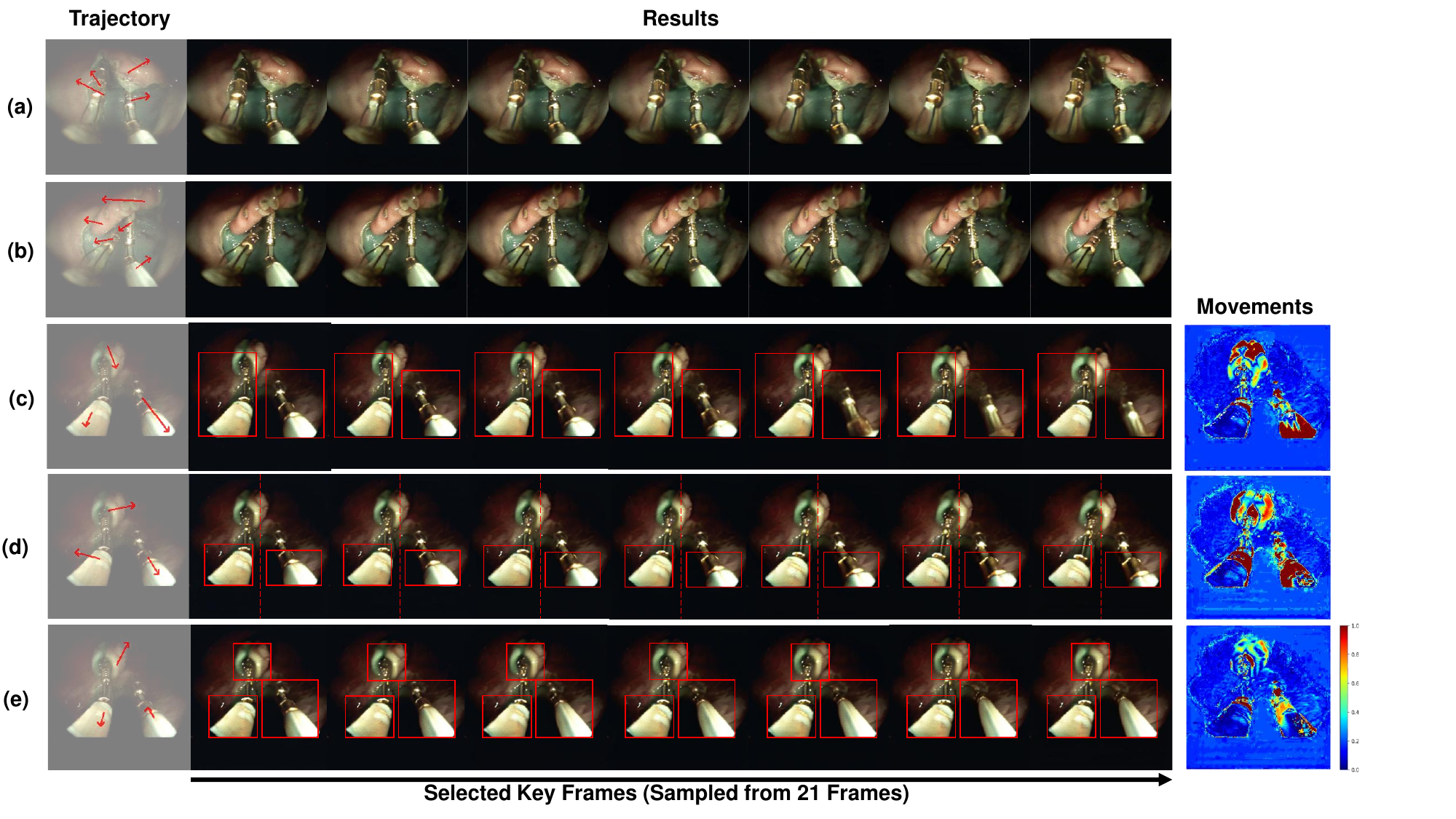}
    \caption{Quantitative results of SurgSora within different trajectories and samples.
    }
    \label{fig: direction}
\end{figure}

Table~\ref{table:baseline} quantitatively assesses SurgSora alongside existing generative models, highlighting our approach's superior performance in video generation. SurgSora achieves the highest frame consistency at 98.70\%, surpassing Endora (97.51\%) and StyleGAN-V (98.02\%), indicating enhanced temporal stability. Besides, it also achieves the best FVD score of 395.65, outperforming StyleGAN-V and Endora, which showcase FVD scores of 857.16 and 1146.54, respectively. Additionally, SurgSora excels in content and dynamics with a CD-FVD score of 535.95, markedly better than MOFA-Video (671.66).
Furthermore, SurgSora leads in visual fidelity with the best FID of 87.94, demonstrating its capacity to generate videos that closely mimic real surgical scenes. The model also tops IS at 3.278, indicating superior object and diversity representation.
\begin{figure}
    \centering
    \includegraphics[width=1\linewidth, trim=-10 30 0 0]{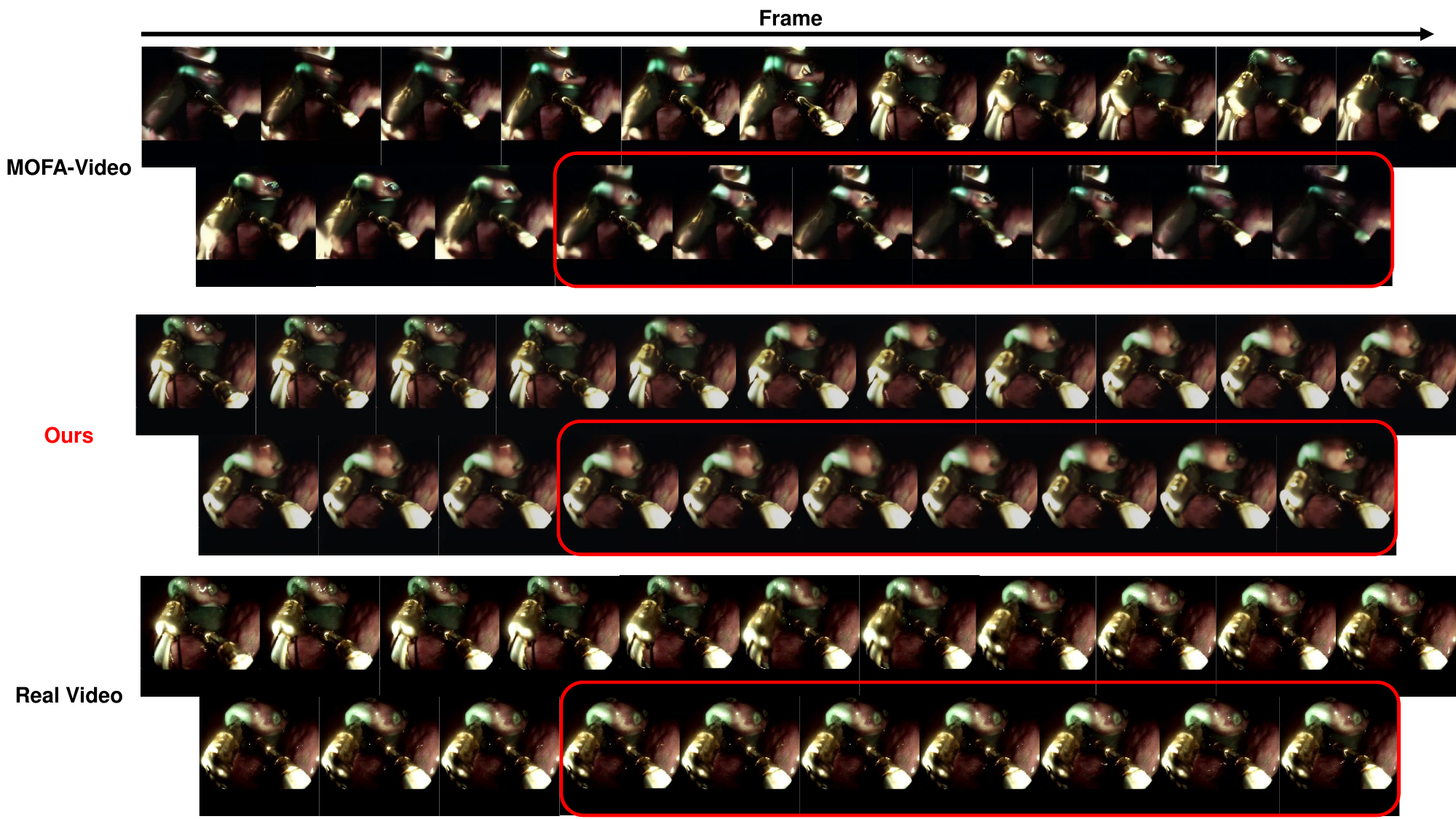}
    \caption{Comparison of conditional generation results trained on CoPESD dataset.}
    \label{fig: baseline1}
\end{figure}
For conditional generation, SurgSora outperforms MOFA-Video with scores of 20.71 dB in PSNR and 55.94\% in SSIM compared to MOFA-Video (19.06 dB and 48.28\%), which confirms promising capacity in preserving generation quality and structural similarity. 
Visual evidence from Fig.~\ref{fig: uncon-baseline} and Fig.~\ref{fig: baseline1} shows SurgSora's superior performance, with videos that maintain authenticity and suffer less distortion compared to other models, showing its practical efficacy in medical applications.
\\
\\
\textbf{Customize Trajectory Video Generation.} To address the effectiveness of our Trajectory Controller block, we generate a few demos by using the TC module. Fig.~\ref{fig: direction} (a)(b) shows videos generated from varying surgical image trajectories, clearly depicting the dynamic movement and transformation of the objects within the images in accordance with the specified trajectories. Further demonstrating the module's capacity for precise control, we generated distinct trajectories within the same image, as presented in Fig.~\ref{fig: direction}(c)(d)(e). We manipulated tissues and instruments to move in designated directions, and the visual results showed that objects along the set paths moved without noticeable distortion. The heatmap indicates that devices have undergone obvious changes according to the trajectory requirements while maintaining the background unchanged. These outcomes validate the high performance and accuracy of our module in controlling and generating detailed movement in medical imagery.
\\
\\
\noindent\textbf{Human Evaluation.} A blind test was conducted to assess the perceptual quality of video clips generated by SurgSora. We randomly selected seven clips from SurgSora-generated videos and eight from real surgical recordings (15 clips in total). These clips were shuffled and anonymized before being presented to six surgeons, who were individually asked to distinguish between generated and real videos. The results, shown in Fig.~\ref{table:baseline}, indicate an average accuracy of 53.33\%, which is close to random guessing (50\%), suggesting that SurgSora produces highly realistic surgical videos that are difficult to differentiate from real footage.
\\
\\
\begin{table}[t]
\centering
	\caption{
        Ablation experiments of our SurgSora on the CoPESD Dataset~\cite{wang2024copesd}.
	}
 	\centering
        \resizebox{\textwidth}{!}{
 \label{tab:ablation}
\begin{tabular}{c|c|c| p{1.3cm}<{\centering} p{1.3cm}<{\centering} p{1.5cm}<{\centering} p{1.3cm}<{\centering} p{1.3cm}<{\centering} p{1.3cm}<{\centering} p{1.3cm}<{\centering}}
\hline
\multicolumn{1}{c|}{\makecell[c]{Segment\\Feature}}  & \multicolumn{1}{c|}{\makecell[c]{Depth 
\\Branch}} & \multicolumn{1}{c|}{\makecell[c]{Multi-Scale\\Fusion}} &\multicolumn{1}{c}{\makecell[c]{Frame 
\\Consistancy}$\uparrow$} &FVD$\downarrow$ &CD-FVD$\downarrow$ & FID$\downarrow$ & IS $\uparrow$ &PSNR$\uparrow$ &SSIM$\uparrow$ \\ \hline
\Checkmark      &\XSolidBrush  &\multicolumn{1}{c|}{\textbf{--}}  &98.08\% &442.66 &584.46 &90.97 &3.199 &20.47 &53.59\%  \\
 \XSolidBrush       & \Checkmark     &\XSolidBrush   &96.99\% &510.11 &782.51 &115.72  &2.586 &19.49 &54.59\% \\
 \XSolidBrush      &\Checkmark      &\Checkmark     &98.35\%  &479.13 &624.63 &88.85 &3.076 &17.59 &53.18\% \\
 \Checkmark       & \Checkmark     &\XSolidBrush   &97.53\% &422.06   &603.34 &88.54 &3.270 &20.64 &51.33\%  \\
\Checkmark      &\Checkmark   &\Checkmark       & \textbf{98.70\%} & \textbf{395.65} & \textbf{535.95}  & \textbf{87.94} & \textbf{3.278}  & \textbf{20.71}&\textbf{55.94\%}\\ \hline
\end{tabular}
}
\end{table}
\noindent\textbf{Ablation Study.} 
We carried out ablation studies on the SurgSora model using the CoPESD Dataset~\cite{wang2024copesd} to assess the impact of different components, summarized in Table \ref{tab:ablation}. Results highlight that removing the segment feature increases FVD/CD-FVD from 395.65/535.95 to 479.13/624.63, underscoring its role in improving visual and temporal coherence. Eliminating the MSF block further degrades performance, with Frame Consistency dropping to 96.99\% and a larger increase in FVD/CD-FVD to 510.11/782.51, emphasizing the depth branch's role in spatial integration). Disabling only the MSF block results in a milder performance drop. The worst structural preservation indicated by the lowest SSIM occurs when decoupled-flow features are used alone, but performance improves with the addition of the depth branch. Integrating all three components yields the best results, confirming their collective utility in video generation quality.

\section{Conclusion}
In this study, we propose SurgSora, a customized RGBD-flow-guided conditional diffusion video model. SurgSora incorporates a separate depth branch, the Dual Semantic Injector (DSI), which increases object semantics information for dual features, and the Decoupled Flow Mapper (DFM) to provide a more suitable and richer feature representation for the Stable Video Diffusion model. Quantitative and qualitative experiments demonstrate superior performance in medical video generation and the ability to generate reasonable videos with simple trajectories. SurgSora provides a brand new view on the medical video generation field. Future works will focus on high-quality long medical clip generation and multimodal conditional medical video generation.

\begin{credits}
\subsubsection{\discintname}
The authors have no competing interests to declare that are relevant to the content of this article.
\end{credits}

\bibliography{Paper-0811}{}
\bibliographystyle{splncs04}
\end{document}